\documentclass[conference]{IEEEtran}
\IEEEoverridecommandlockouts

\usepackage{cite}
\usepackage{amssymb}
\usepackage{amsmath}
\usepackage{graphicx}
\usepackage{subfigure}
\usepackage{pifont}
\usepackage{comment}
\usepackage{color}
\usepackage[strings]{underscore}
\PassOptionsToPackage{hyphens}{url}
\usepackage{url}

\renewcommand{\vec}[1]{\mathbf{#1}} %
\newcommand{\vecg}[1]{\boldsymbol{#1}} %
\newcommand{\red}{}
\newcommand{\rev}{}
\newcommand{\blue}{}

\newcommand{\sect}{\textsection}

\begin{document}
\title{Sardino: Ultra-Fast Dynamic Ensemble for Secure Visual Sensing at Mobile Edge}
\author{\IEEEauthorblockN{Qun Song\IEEEauthorrefmark{1} $\qquad$
    Zhenyu Yan\IEEEauthorrefmark{2} $\qquad$
    Wenjie Luo\IEEEauthorrefmark{1} $\qquad$
    Rui Tan\IEEEauthorrefmark{1}}
  \IEEEauthorblockA{
    \IEEEauthorrefmark{1}\textit{Nanyang Technological University, Singapore}
    $\quad$
    \IEEEauthorrefmark{2}\textit{Chinese University of Hong Kong, HKSAR, China}}
}

\maketitle

\thispagestyle{plain}
\pagestyle{plain}

\begin{abstract}
  Adversarial example attack endangers the mobile edge systems such as vehicles and drones that adopt deep neural networks for visual sensing. This paper presents {\em Sardino}, an active and dynamic defense approach that renews the inference ensemble at run time to develop security against the adaptive adversary who tries to exfiltrate the ensemble and construct the corresponding effective adversarial examples. By applying consistency check and data fusion on the ensemble's predictions, Sardino can detect and thwart adversarial inputs. Compared with the training-based ensemble renewal, we use HyperNet to achieve {\em one million times} acceleration and per-frame ensemble renewal that presents the highest level of difficulty to the prerequisite exfiltration attacks. We design a run-time planner that maximizes the ensemble size in favor of security while maintaining the processing frame rate. Beyond adversarial examples, Sardino can also address the issue of out-of-distribution inputs effectively. This paper presents extensive evaluation of Sardino's performance in counteracting adversarial examples and applies it to build a real-time car-borne traffic sign recognition system. Live on-road tests show the built system's effectiveness in maintaining frame rate and detecting out-of-distribution inputs due to the false positives of a preceding YOLO-based traffic sign detector.
\end{abstract}

\section{Introduction}
\label{sec:intro}

Deep neural network (DNN)-based visual sensing is an important perception approach for mobile edge systems such as vehicles and drones. In Apollo \cite{apollo}, which is an autonomous vehicle driving agent, the tasks of recognizing road signs, traffic lights, and lane markers are implemented with DNN-based visual sensing.
However, the {\em adversarial example} attack introduces much concern. Recent research shows that an external adversary can systematically craft minute perturbations added to the inference samples and mislead a DNN to yield absurd results \cite{goodfellow6572explaining}. 
Readily deployable adversarial examples like paper stickers pasted on the road \cite{tencent} and traffic sign plate \cite{eykholt2018robust} are shown effective against lane detection and traffic sign recognition systems. Thus, the designs of DNN-based visual sensing for safety-critical mobile edge systems should incorporate effective defense against adversarial examples.

Various countermeasures have been proposed, e.g., adversarial training \cite{goodfellow6572explaining}, input transformation \cite{ren2020adversarial}, gradient masking \cite{athalye2018obfuscated}, and provable defenses \cite{wong2018provable}. These approaches build their security upon the assumption that the adversary is ignorant of the defense mechanisms. Such {\em static} defenses can be breached if the adaptive adversary obtains the details of the defense mechanisms and designs the next-generation attacks \cite{athalye2018obfuscated,ren2020adversarial}.
Using an ensemble of multiple distinct DNNs has also been considered as a defense \cite{he2017adversarial}. Specifically, the ensemble uses some rule (e.g., majority vote) to combine multiple DNNs' inference results to generate a final result. Intuitively, it becomes harder for an adversarial example to mislead multiple DNNs than a single DNN. However, the adaptive adversary who has exfiltrated the ensemble can subvert the ensemble-based defense with substantial probabilities (e.g., 52\% as shown in \sect\ref{sec:measurement-study} of this paper). The static ensemble can be exfiltrated from the mobile edge device's memory or by social engineering against the system designer's employees.

To strengthen mobile edge's visual sensing security against adaptive adversary, we propose using {\em dynamic ensemble} for active defense under the strategy of {\em moving target defense} (MTD) \cite{jajodia2011moving}. MTD improves system security and increases the difficulties for effective attacks by dynamically changing the system configurations at run time.
In this paper, the ensemble is renewed frequently at run time and unpredictable by the adversary.
This approach's effectiveness stems from a basic observation that the adversarial examples have limited transferability to the DNNs different from those used for attack construction.
Its security strength is greatly affected by the following two aspects. First, larger ensemble sizes and higher renewal rates enhance security strength. Specifically, it is harder for the adversary to construct adversarial examples that can mislead all DNNs of a larger ensemble. Meanwhile, if the ensemble is renewed more frequently, the adaptive adversary has shorter time for exfiltrating the ensemble. Second, higher diversity of an ensemble's DNNs fosters attack detection, because these DNNs tend to produce more diverse classification results for an adversarial example input.

An earlier study \cite{song2019moving} uses dynamic ensembles to counteract adaptive adversarial example attacks. It is based on a primitive approach of retraining DNNs using data stored on the mobile edge device,
which impedes achieving high-rate ensemble renewal.
As shown in \cite{song2019moving}, it takes 45 minutes on NVIDIA Jetson AGX Xavier to retrain an ensemble of 20 DNNs for a traffic sign classification task. As the retraining process is very compute-intensive,
the ensemble renewal in \cite{song2019moving} is performed when the mobile system is idle (e.g., when the car is parked). In the cases of fuel cars, it requires using the car battery to power the lengthy retraining. If multiple task ensembles are renewed, the retraining risks battery over-discharge. In addition, the retraining requires a large training dataset stored on the mobile edge, which is cumbersome.

\begin{figure*}
  \centering
  \includegraphics[width=.95\textwidth]{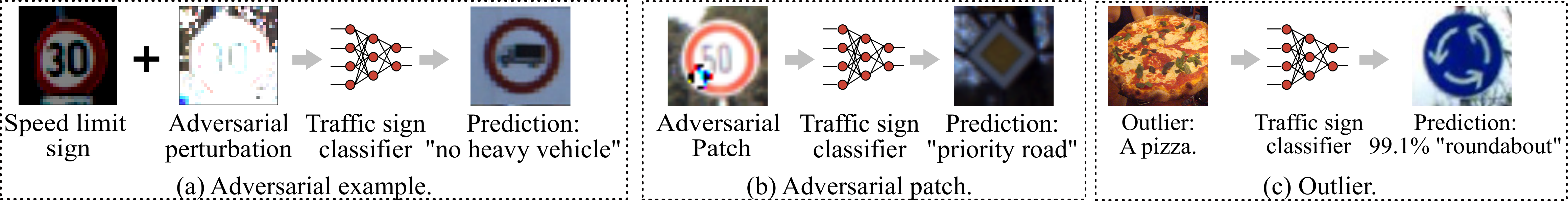}
  \vspace{-1em}
  \caption{Illustration of the adversarial example and OOD input challenges for DNN-based visual sensing.}
  \label{fig:outlier-adv-vs-pred}
  \vspace{-1em}
\end{figure*}

In contrast to the off-time, infrequent ensemble renewal achieved in \cite{song2019moving}, this paper aims to achieve run-time and high-rate ensemble renewal, which gives two advantages. First, higher renewal rates mean better MTD security. Second, in the context of cars and drones, run-time renewal avoids lengthy battery discharge during parking. However, the run-time renewal and the execution of large ensembles in favor of security should be precisely managed to avoid jeopardizing the real-time performance of the visual sensing. In this paper, we design {\em Sardino}\footnote{Sardino is the Esperanto word of sardine. When threatened, sardines form a school that undertakes complicated maneuvers and startling shape changes. The many moving targets of the school create a sensory overload of the predator's visual and electrosensory channels \cite{milinski1978influence}.} to achieve the goal. Specifically, the design of Sardino consists of the following two main aspects.

First, we follow the {\em HyperNet} concept \cite{ratzlaff2019hypergan} to design the DNN generator for fast ensemble renewal. The generator is a set of multilayer perceptrons (MLPs) that take random numbers as input and generate the weights of DNNs. A key advantage of Sardino is that the ensemble renewal becomes forwarding the MLPs, which is much faster than DNN training and does not require storing training data on the mobile.
We show that generating a DNN for the aforementioned traffic sign classification task on Jetson AGX Xavier only takes 0.1 milliseconds, which is 0.66 and 1.35 million times faster than the two DNN retraining approaches in \cite{song2019moving,motiian2017few}.
Owing to the accelerated DNN generation, Sardino achieves per-frame ensemble renewal that renders the highest MTD security.


Second, we design a run-time ensemble size planner, such that the total delay of renewing and executing the ensemble on a mobile edge device shared by other continuing inference tasks meets a soft deadline determined by the sensing frame rate. To this end, the ability to predict the delay is needed but developing this ability is non-trivial. With extensive profiling experiments, we identify that the latest GPU utilization and power usage are two factors affecting the delay. With a decision tree regressor that predicts the delay based on the affecting factors,
we maximize in real time the ensemble size in favor of security, subject to the deadline. In other words, Sardino uses the available compute time to increase security. 

Adversarial examples can be viewed as a crafted type of out-of-distribution (OOD) inputs that fall out of the training data distribution. In practice, naturally occurring OOD inputs are common. Since Sardino can address adversarial examples under a highly adversarial setting, it can also address the naturally occurring OOD inputs. To demonstrate this, we implement a real-time car-borne traffic sign recognition system based on Sardino.
Extensive evaluation including live on-road tests shows the effectiveness of Sardino in meeting soft deadlines and detecting OOD inputs due to the preceding YOLO's \cite{yolo} false positives in detecting traffic signs.

This paper's main contributions are summarized as follows:
\begin{itemize}
\item We propose Sardino for high-rate ensemble renewal to defeat the external adversary's DNN exfiltration as a prerequisite for adversarial example construction. We design a HyperNet to implement the high-rate renewal.
\item
  Extensive evaluation shows Sardino's superior performance in counteracting both adversarial examples and naturally occurring OOD inputs, compared with the existing retraining approaches \cite{song2019moving,motiian2017few} and the HyperGAN approach \cite{ratzlaff2019hypergan}.
\item We design an ensemble size planner to meet a specified soft deadline for ensemble renewal and execution, which is imperative to real-time visual sensing. The design is applicable to the execution on either graphics processing unit (GPU) or central processing unit (CPU).
\end{itemize}

{\em Paper organization:} \sect\ref{sec:background} presents background. \sect\ref{sec:overview-of-ultramtd} overviews Sardino. \sect\ref{sec:measurement-study} studies effectiveness of dynamic ensemble. \sect\ref{sec:ensemble-cardinality-scheduling} presents the ensemble size planner. \sect\ref{sec:traffic-sign-recognition} presents the car-borne traffic sign recognition system. \sect\ref{sec:discuss} discusses several related issues. \sect\ref{sec:conclusion} concludes this paper.
\section{Background}
\label{sec:background}

\subsection{\blue Adversarial Examples and OOD Data}
\label{subsec:challenges}

Consider a classifier $f(\cdot; \vecg{\theta})$ with weights $\vecg{\theta}$ that classifies an input $\vec{x}$ as $y$, i.e., $f(\vec{x}; \vecg{\theta}) = y$. An adversarial example $\vec{x}' = \vec{x} + \vecg{\delta}$, where $\vecg{\delta}$ is a perturbation, results in $f(\vec{x}'; \vecg{\theta}) \neq y$. The magnitude of $\vecg{\delta}$ is often minimized to reduce perceptual change. Fig.~\ref{fig:outlier-adv-vs-pred}a and Fig.~\ref{fig:outlier-adv-vs-pred}b illustrate the impacts of adversarial examples on a convolutional neural network (CNN) trained for traffic sign recognition. In Fig.~\ref{fig:outlier-adv-vs-pred}a, the $\vecg{\delta}$ is computed by the Carlini and Wagner (C\&W) method \cite{carlini2017towards} and added to a clean speed limit sign, leading to a wrong classification of ``no heavy vehicle.'' When the adversary cannot tamper with each pixel, they may construct {\em adversarial patches} \cite{brown2017adversarial}. In Fig.~\ref{fig:outlier-adv-vs-pred}b, an adversarial patch \cite{brown2017adversarial} is added to a speed limit sign, leading to a wrong classification of ``priority road.'' Such adversarial patches pasted on road can mislead Tesla Autopilot to direct a car to the opposite lane \cite{tencent}, creating great danger.

Besides adversarial examples, OOD data or outliers can naturally occur because a training dataset cannot include all future unseen data \cite{hendrycks2016baseline}. A DNN may have high confidence about its wrong classification for an OOD input. As illustrated in Fig.~\ref{fig:outlier-adv-vs-pred}c, when the input is a pizza picture, the aforementioned CNN
yields a ``roundabout'' classification result with a high confidence score of 99.1\%.
Tesla Autopilot has made similar mistakes, e.g., recognize a Burger King sign as a stop sign \cite{tesla-1} and moon as yellow traffic light \cite{tesla-2}. Although the CNN in Fig.~\ref{fig:outlier-adv-vs-pred}c can be retrained to recognize pizzas by adding pizza image samples to the training dataset, this approach cannot cover every possible non-traffic sign object.

This paper aims to improve the resilience of the mobile edge's visual sensing against the issues illustrated in Fig.~\ref{fig:outlier-adv-vs-pred}.

\subsection{Related Work}
\label{subsec:related}

The existing countermeasures against adversarial examples are categorized as follows \cite{ren2020adversarial}. {\em Adversarial training} \cite{goodfellow6572explaining} includes adversarial examples in the training dataset. The enhanced DNN is secure against the adversarial examples considered during adversarial training. 
{\em Transformations} 
on input 
such as random resizing/padding, image compression, and noisification 
are shown effective against the attack. However, they can be defeated by attackers who know the adopted transformation \cite{ren2020adversarial}.
{\em Gradient masking} \cite{athalye2018obfuscated} manipulates the victim DNN's gradients to render gradient-based attacks ineffective. However, an adversary aware of the defense can recover the gradients by querying the victim DNN or use other loss functions to construct attack \cite{athalye2018obfuscated}.
{\em Provable defense} \cite{wong2018provable} develops certifiable methods that give lower-bounded defense effectiveness against a certain class of attacks.
The above defenses cannot address adaptive adversary. Sardino addresses such adaptive adversary by updating the ensemble at a speed faster than the adversary's exfiltration for the ensemble.

A method to detect outliers is to train DNNs to make highly uncertain predictions for outliers \cite{hendrycks2018deep}.
The work in \cite{ruff2018deep} trains a one-class neural network to detect outliers. 
The study in \cite{ren2019likelihood} trains a generative model and evaluates the likelihood of OOD inputs under that model at run time.
Static ensembles generated by HyperNet \cite{ratzlaff2019hypergan} or other methods \cite{he2017adversarial} have been used to counteract outliers and adversarial examples. Their focuses are on the trade-off between the ensemble size and OOD/attack detection performance, under the non-adaptive adversary setting. As shown in \sect\ref{sec:measurement-study}, the adversary can construct effective adversarial examples once they obtain the ensemble. The work \cite{ratzlaff2019hypergan} does not exploit the key advantage of HyperNet, i.e., its ability to renew the ensemble quickly for implementing MTD.

Recent studies aim to improve GPU utilization and processing throughput under a multi-tasking setting.
The work \cite{zhou2018s} schedules multiple DNN tasks in the granularity of GPU kernels for improved GPU utilization. The work \cite{narayanan2018accelerating} achieves acceleration by performing operator fusion and I/O sharing across multiple DNNs. 
In above studies, the GPU is accessed by the kernels in a round-robin fashion. 
Alternatively, multiple kernels can run simultaneously on their own GPU cores to enable {\em spatial sharing}.
The work \cite{zhang2018g} uses this to improve GPU-accelerated network function virtualization.
The above studies \cite{zhou2018s,narayanan2018accelerating,zhang2018g} 
focus on task scheduling to maximize processing throughput and do not enforce deadlines. Differently, Sardino plans the ensemble size at run time, aiming at meeting a soft deadline for using the ensemble to process each frame.

\section{Approach Overview}
\label{sec:overview-of-ultramtd}

\subsection{System Model and Objective}
\label{subsec:system-model}

We consider a mobile edge computer equipped with a GPU.
It runs a general-purpose operating system that orchestrates various sensing tasks.
The GPU is shared by the tasks for executing their DNNs. The tasks run simultaneously, take inputs from respective sensors, and yield the inference results. Among all the tasks, we focus on a {\em resilient vision task} that needs to have resilience against adversarial examples and OOD inputs, and meet a soft deadline. We view the composite of the remaining tasks as the {\em background computation}. 

We apply dynamic ensemble for the resilient vision task. The ensemble is dynamic in both the number of DNNs of the ensemble (i.e., ensemble size) and the weights of each DNN.

\begin{LaTeXdescription}
\item[Dynamic ensemble size:] We aim to maximize in real time the ensemble size for every frame in favor of resilience subject to the soft deadline. However, it is challenging to model the ensemble execution time to enable the run-time ensemble size planning.
\item[Dynamic DNN weights:] We also aim to renew the weights of each DNN every frame for achieving the highest level of MTD security. The short time of down to milliseconds for completing the renewal presents the main challenge.
\end{LaTeXdescription}

We use traffic sign recognition on an autonomous vehicle driving agent to illustrate the system model. The agent receives image frames captured by camera and stores them in a buffer. Each frame fetched from the buffer is processed by an always-running traffic sign detector. When the detector identifies $k$ ($k \ge 1$) traffic sign objects in the current frame, a bounding box containing each of the detected traffic signs is cropped from the frame and passed to the traffic sign classifier. The classifier is executed on each detected sign sequentially. The detected traffic sign may contain adversarial perturbations \cite{eykholt2018robust}. Moreover, the traffic sign detector may generate false positives and present outliers to the traffic sign classifier. We view the classifier as resilient vision task and all other tasks collectively as background computation.
Suppose the system designer aims to maintain a processing throughput of $x$ frames per second (fps). Thus, the soft deadline for processing each frame is $\frac{1}{x}$ seconds. If the detection takes $t_d$ seconds for the current frame containing $k$ signs, the soft deadline for classifying each sign is $\frac{1/x - t_d}{k}$ seconds. Although $t_d$ and $k$ vary across the frames, they are measurable. Thus, the soft deadline for the classifier, i.e., $\frac{1/x - t_d}{k}$, is variable and known for each frame. The setting for $x$ depends on the vision task's design requirement; it can be also updated at run time according to the vehicle speed.

In \sect\ref{sec:traffic-sign-recognition}, we will apply Sardino to implement the resilient traffic sign recognition. {\red In addition, \sect\ref{sec:discuss} will discuss how to apply Sardino to protect both the traffic sign detection and recognition simultaneously as two resilient vision tasks.}

\subsection{HyperNet Design and Its Adversarial Learning}
\label{subsubsec:HyperNet-design}

\begin{figure}
  \includegraphics[width=\columnwidth]{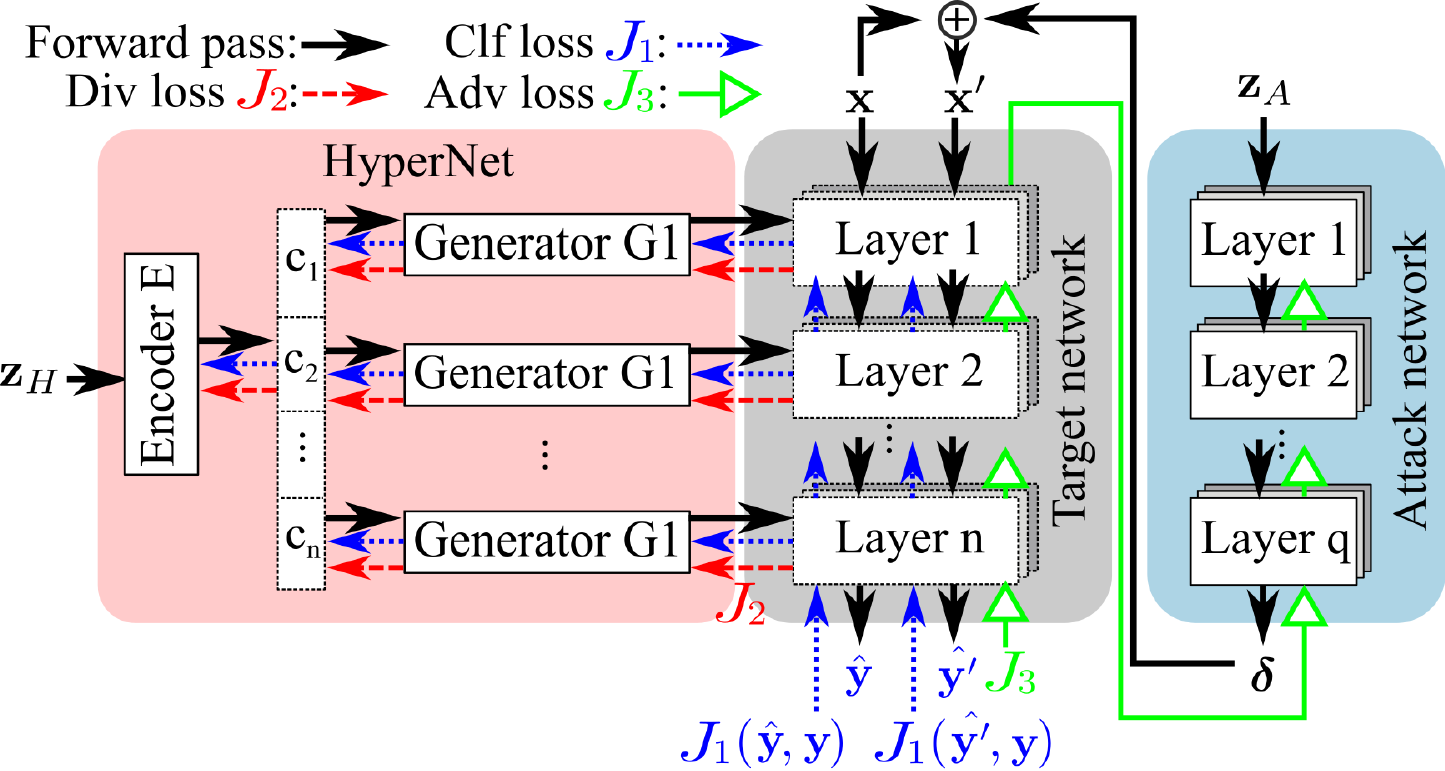}
  \vspace{-2em}
  \caption{Adversarial learning framework for Sardino.}
  \vspace{-1em}
  \label{fig:hypernet-arch}
\end{figure}

\subsubsection{Hypernet preliminaries}
HyperNet \cite{ratzlaff2019hypergan} is a neural network (denoted by $h(\cdot; \vecg{\phi})$ where $\vecg{\phi}$ denotes the weights) that generates the weights $\vecg{\theta}$ of the {\em target neural network} denoted by $f(\cdot; \vecg{\theta})$.
Fig.~\ref{fig:hypernet-arch} shows the designs of the HyperNet and target network, which is a $n$-layer CNN, used in this paper.
The input to the HyperNet, denoted by $\vec{z}_H$, is a random vector sampled from a normal distribution.
The $\vec{z}_H$ is mapped by an {\em encoder} $E$ with weights $\vecg{\phi}_E$ to $n$ {\em latent codes} $\{c_i | i = 1, 2, \ldots, n\}$. Then, the HyperNet uses $n$ {\em weight generators} with weights $\vecg{\phi}_G$ to convert the latent codes to the weights of the target CNN's $n$ layers. The HyperNet's weights are $\vecg{\phi} = \{\vecg{\phi}_E, \vecg{\phi}_G\}$.

\subsubsection{Adversarial learning framework}
\label{sec:adv-ml}

In this paper, the goal for training the HyperNet is to generate target networks that are: (1) accurate on clean input samples; (2) diverse in parameter values; and (3) secure against adversarial examples. This subsection presents the designs of the loss functions and the training procedure to meet the three objectives. 

For objective (1), we define the {\em classification loss} (denoted by $J_1$) by the average cross-entropy loss on input examples:
\begin{equation*}
  J_1 = L(f(\vec{x}; G(E(\vec{z}_H;\vecg{\phi}_E); \vecg{\phi}_G)),y),
\end{equation*}
where $E(\vec{z}_H;\vecg{\phi}_E)$ represents latent code, $G(E(\vec{z}_H;\vecg{\phi}_E); \vecg{\phi}_G)$ denotes target network's weights generated by the HyperNet, $f(\vec{x}; G(E(\vec{z}_H;\vecg{\phi}_E); \vecg{\phi}_G))$ is the target network's classification result for input $\vec{x}$, and $L(\cdot, \cdot)$ denotes cross-entropy.
For objective (2), we design the {\em diversity loss} denoted by $J_2$ as:
\begin{equation*}
  J_2 = \exp(- \mathrm{Var}(G(E(\vec{z}_H;\vecg{\phi}_E); \vecg{\phi}_G))),
\end{equation*}
where $\mathrm{Var}(\cdot)$ is the average variance of the generated target network's weights given a batch of $\vec{z}_H$. As $\mathrm{Var}(\cdot)$ is not bounded, we apply the exponential function to avoid divergence.
For objective (3), {\rev inspired by \cite{primask},} we employ the {\em adversarial learning} technique \cite{huang2011adversarial} {\rev to train the HyperNet}. Adversarial learning addresses a game between a {\em defender} that trains a task model to thwart the attacker's objective and an {\em attacker} that trains an attack model to mislead the defender's task model. {\rev In \cite{primask}, an {\em attack network} is designed to be the attacker that tries to breach the privacy protection mechanism provided by the HyperNet-based defender. In this work, we design an attack network, as shown in Fig.~\ref{fig:hypernet-arch}, to be the attacker that tries to generate adversarial examples to mislead the HyperNet-generated target network.}
The attack network $f_A(\cdot; \vecg{\theta}_A)$ takes random numbers $\vec{z}_A$ sampled from a normal distribution and outputs adversarial perturbation $\vecg{\delta}$. The perturbation $\vecg{\delta}$ is added to the clean input $\vec{x}$, forming the adversarial example $\vec{x}'$. The goal of training the attack network is to generate minimized adversarial perturbations that mislead the target network $f(\cdot; \vecg{\theta})$. 
Thus, the {\em adversarial loss} for training the attack network, denoted by $J_3$, is designed as:
\begin{equation*}
  J_3 = F(\vec{x}')_y - \max_{y_i \neq y} F(\vec{x}')_{y_i} + \| \vecg{\delta}\|_2,
\end{equation*}
where $F(\cdot)_{y_i}$ denotes the target network's logit value corresponding to class $y_i$. Logit value is the output of neural network's last layer before applying the softmax function. $\|\cdot\|_2$ is the Euclidean norm. 
The attack network and HyperNet are jointly trained, where the attack network is trained to minimize the loss $J_3$ and the HyperNet is trained to minimize the following composite loss: $\mathbb{E}_{\vec{z}_H,\vec{z}_A,(\vec{x},y)} [J_1] + \mathbb{E}_{\vec{z}_H} [J_2]$.
Fig.~\ref{fig:hypernet-arch} illustrates the training procedure of the adversarial learning. 

During the adversarial learning, we do not employ specific methods for crafting adversarial examples (e.g., FGSM, C\&W), because doing so usually leads to security improvement specific to the employed attack construction methods only \cite{ren2020adversarial}. However, in reality, the attack construction method is unpredictable. Our design uses the attack network to generate nondeterministic adversarial examples, which improves Sardino's security against a variety of adversarial examples. This will be demonstrated in \sect\ref{sec:thwarting-adversarial}.

\subsection{\blue System Design of Sardino}
\label{subsec:UltraMTD-workflow}

\begin{figure}
  \centering
  \includegraphics[width=\columnwidth]{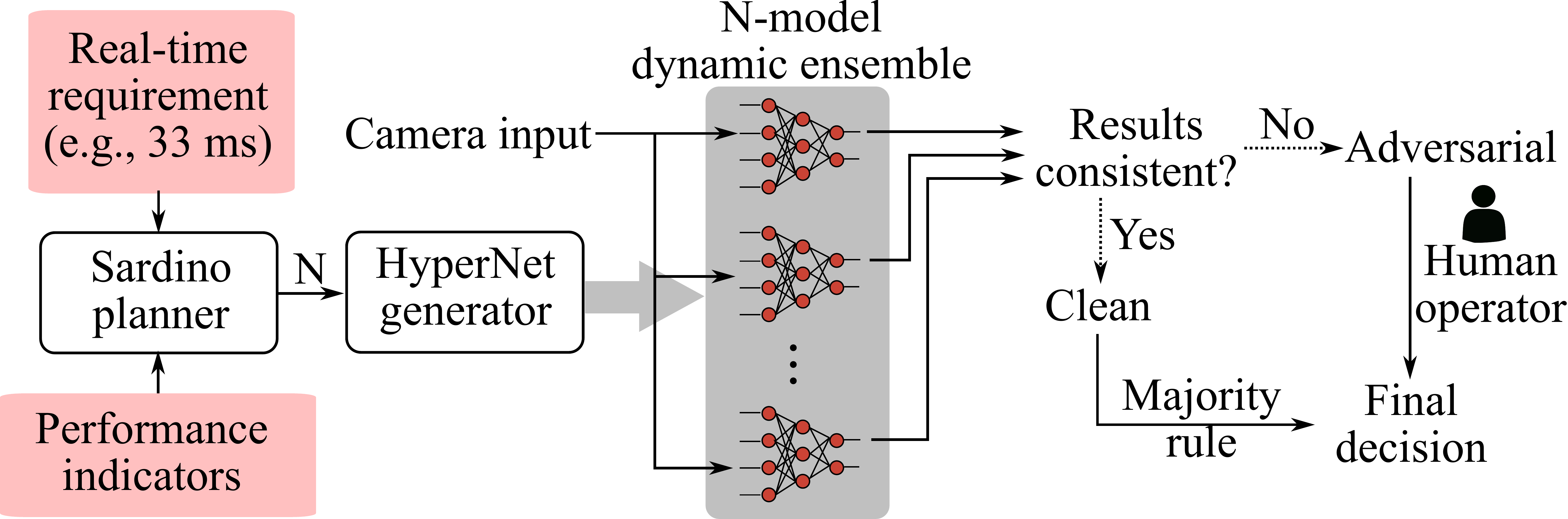}
  \vspace{-2em}
  \caption{Run-time workflow of Sardino.}
  \label{fig:approach-overview}
  \vspace{-1em}
\end{figure}

{\blue 
  Fig.~\ref{fig:approach-overview} illustrates the run-time workflow of Sardino. Given a new image frame, Sardino uses the HyperNet to generate a new ensemble of $N$ DNNs to process the input. Before the generation, Sardino uses an ensemble size planner to determine the largest possible $N$ based on the mobile edge device's performance indicators (i.e., GPU utilization and power usage) and the soft deadline described in \sect\ref{subsec:system-model}.
  After the execution of the $N$ DNNs on the image frame, Sardino computes the \textit{output consistency}, which is the percentage of the majority of the DNNs' outputs. If the consistency is larger than a pre-defined threshold $T_s$, the input is considered clean and the majority of the DNNs' outputs is yielded as the final result. If the output consistency is smaller than $T_s$, the input is considered adversarial or OOD, and will be classified by a human operator for final decision.
In summary, based on the mobile edge device's run-time performance indicators, Sardino adapts the ensemble size $N$ to meet the soft real-time requirement, then generates and executes the dynamic ensemble to process the incoming image frame.
}
\section{Effectiveness of Dynamic Ensemble}
\label{sec:measurement-study}

\subsection{Profiling Experiment Setup}
\label{sec:measurement-setup}

We conduct experiments on NVIDIA Jetson AGX Xavier with an octa-core 2.26GHz ARM CPU, a 512-core Volta GPU, and 16GB RAM. We set its power rating to be 30W. It runs Linux4Tegra. We write code in Python using PyTorch 1.4.0.

Most experiments are based on German Traffic Sign Recognition Benchmark (GTSRB) dataset \cite{stallkamp2011german} with over 50,000 image samples in 43 classes.
To evaluate outlier detection, we use the MNIST \cite{mnist} and notMNIST \cite{bulatov2011notmnist} datasets. MNIST is a 10-class set of grayscale images of handwritten digits from 0 to 9; notMNIST is a 10-class set of grayscale images of letters A to J, which is often used for studying outlier detection \cite{ratzlaff2019hypergan}.

The target CNN has two convolutional layers with 32 5x5 filters with rectified linear unit activation, max pooling, and a dense layer with width equal to the class number. HyperNet's encoder has two 64-neuron dense layers and a dense layer with 64x3 neurons. The encoder's input is a 256x1 Gaussian random vector. The encoder is followed by three weight generators, each of which has two 64-neuron dense layers and one dense layer with identical width as the output layer.

\subsection{Profiling Experiments and Results}
\label{sec:measurement-results}

\subsubsection{Classification performance of HyperNet ensemble}
\label{sec:classification-performance}

\begin{figure}
  \centering
  \begin{minipage}{.48\columnwidth}
      \centering
      \includegraphics[width=.85\columnwidth]{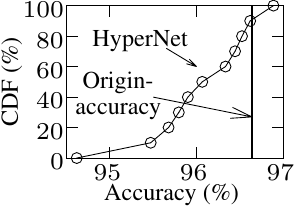}
      \vspace{-1em}
      \caption{CDF of {\blue 100,000} HyperNet-generated DNNs' accuracy on GTSRB dataset.}
      \label{fig:test_acc_N_100_CDF}
  \end{minipage}
  \hfill
  \begin{minipage}{.48\columnwidth}
    \centering
    \includegraphics[width=.85\columnwidth]{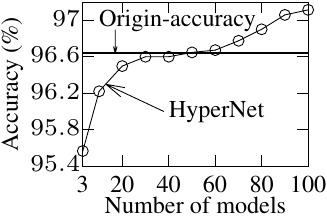}
    \vspace{-0.5em}
    \caption{Test accuracy vs. size of HyperNet-generated ensemble (Dataset: GTSRB).}
    \label{fig:gtsrb-acc}
  \end{minipage}
  \vspace{-1em}
\end{figure}

The curve in Fig.~\ref{fig:test_acc_N_100_CDF} is the cumulative distribution function (CDF) of the test accuracies of {\blue 100,000} HyperNet-generated DNNs. The vertical line labeled {\em origin-accuracy} is the test accuracy of the original DNN trained from the GTSRB dataset following the design in \cite{stallkamp2011german}, which is 96.6\%.
The accuracies of HyperNet-generated DNNs are within 94.6\% to 96.9\%, 
showing that HyperNet can generate quality DNNs. 
We also investigate the accuracy improvements of fusing the outputs of multiple HyperNet-generated DNNs using the majority rule (called HyperNet ensemble). As shown in Fig.~\ref{fig:gtsrb-acc}, the HyperNet ensemble's accuracy increases with ensemble size $N$.
In particular, compared with the accuracy when no fusion is applied, the accuracy improvement is up to 1\% when $N$ is 3. When $N$ is 100, the improvement is up to 2.5\% and the ensemble's accuracy is 0.5\% higher than the origin-accuracy.

\subsubsection{Performance in thwarting adversarial examples}
\label{sec:thwarting-adversarial}

The key objective of dynamic ensemble is to prevent the external adversary from obtaining the ensemble in use. 
Our test shows that, if the adversary obtains the ensemble in use, the adversarial example constructed by the approach in \cite{liu2016delving} can mislead the ensemble-based attack detection described in \sect\ref{subsec:UltraMTD-workflow} with probabilities of 52\% and 37\% when the false positive rates are 1.1\% and 5.2\%, respectively. Thus, high-rate ensemble renewal is key to MTD security. 
Note that the internal adversary who has broken into the system and can obtain each renewed ensemble regardless of renewal rate is out of the scope of this paper, since the internal adversary should directly subvert the whole system rather than resort to adversarial examples.
In this set of profiling experiments, we study a different external adversary who obtains some critical static information about the dynamic ensemble. We consider two kinds of static information: (1) training dataset and (2) the HyperNet itself.
We evaluate the attack thwarting performance for five variants of the ensemble-based detector, namely, {\em retraining-ensemble} from \cite{song2019moving}, {\em few-shot retraining-ensemble} from \cite{motiian2017few}, {\em HyperGAN-ensemble} from \cite{ratzlaff2019hypergan}, and {\em HyperNet-ensemble} proposed in this paper with and without adversarial learning. 
In {\em retraining-ensemble}, each DNN is trained from scratch with random initialization.
Each DNN of the {\em few-shot retraining-ensemble} is obtained by the few-shot domain adaptation approach in \cite{motiian2017few} that adapts a base model trained with a big source-domain data subset to the target domain using a small data subset containing 7 samples for each class.
{\em HyperGAN-ensemble} is generated by HyperGAN \cite{ratzlaff2019hypergan} that trains a generator to transform random numbers into target network's weights together with the help of a discriminator to promote the diversity of the generated weights.

\begin{figure*}
  \centering
  \includegraphics[width=.9\textwidth]{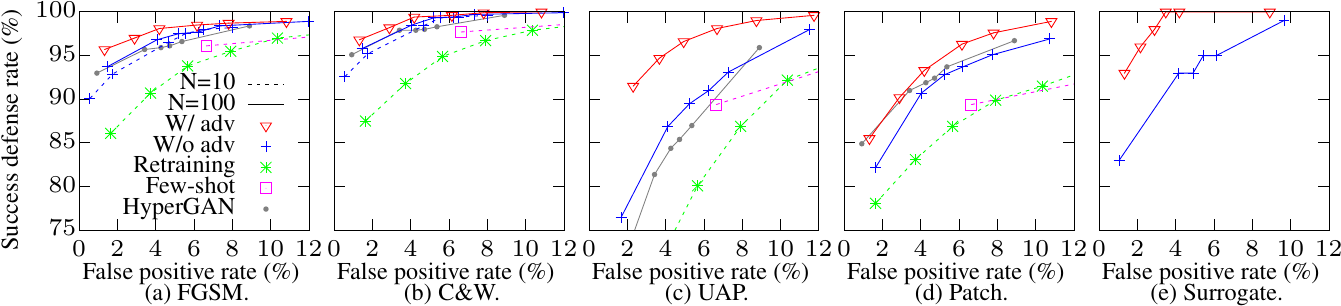}
  \vspace{-1em}
  \caption{SDR vs. FPR in thwarting various types of adversarial examples, i.e., FGSM \cite{goodfellow6572explaining} in (a), C\&W \cite{carlini2017towards} in (b), UAP \cite{moosavi2017universal} in (c), Patch \cite{brown2017adversarial} in (d), and that against surrogate ensemble \cite{liu2016delving} in (e). The lines labeled with "W/ adv" and "W/o adv" are the results of {\em HyperNet-ensemble} with and without adversarial learning. Legends for (b)-(e) are same as (a).}
  \label{fig:sdr-vs-fpr-gtsrb-seperate-attack}
  \vspace{-1em}
\end{figure*}

$\blacksquare$ {\bf Adversary with training dataset:} 
The adversary can train a {\em surrogate DNN} and use different methods to construct adversarial examples against it. There are two types of adversarial examples \cite{ren2020adversarial}:
{\em input-specific} perturbation is crafted against a specific clean sample, while ideally, {\em universal} perturbation is effective against any clean sample.
In this paper, we consider two input-specific attacks, which are FGSM \cite{goodfellow6572explaining} and C\&W \cite{carlini2017towards}, and two universal attacks, which are universal adversarial perturbation (UAP) \cite{moosavi2017universal} and adversarial patch (Patch) \cite{brown2017adversarial}.
From our measurements, the FGSM, C\&W, UAP and Patch attacks can mislead the surrogate DNN on 97.4\%, 100\%, 45\% and 33.1\% of clean test samples.
We follow the workflow in Fig.~\ref{fig:approach-overview} with $N$ fixed to implement the defense. 
We assume that the detected adversarial examples are classified by the human operator without errors, since adversarial perturbations are crafted to be visually imperceptible.
We measure the {\em successful defense rate} (SDR), which is the percentage of the adversarial examples failing to mislead the system.

Fig.~\ref{fig:sdr-vs-fpr-gtsrb-seperate-attack} shows SDR versus the false positive rate (FPR) in thwarting adversarial examples, where FPR characterizes the unnecessary overhead incurred to the human operator. 
From Fig.~\ref{fig:sdr-vs-fpr-gtsrb-seperate-attack}a-d, {\em HyperNet-ensemble} with adversarial learning produces the highest curves, i.e., the best trade-off between the security and the overhead incurred to human. An intuitive explanation for the better attack thwarting performance of the {\em HyperNet-ensemble} over the {\em HyperGAN-ensemble} is that HyperGAN only increases the diversity of the generated weights and does not consider adversarial examples during training.
When FPR is around 2\%, SDRs of the {\em HyperNet-ensemble} with adversarial learning are 96.3\%, 97.5\%, 91.5\%, and 88.2\% against the four attacks, respectively. For a certain attack, when $N$ increases, the curve becomes higher. This indicates that larger $N$ settings are beneficial to the effectiveness of defense.

We also compare our approach with an adversarial training approach \cite{madry2018towards} in terms of defense performance. The adversarial training approach includes adversarial examples constructed using the project gradient descent method \cite{kurakin2017adversarial} into the training dataset.
It achieves SDRs of 35.7\%, 25\%, 85\%, and 68\% against the four attacks.
Its poor defense is due to that adversarial training's effectiveness is specific to the considered type of adversarial examples \cite{ren2020adversarial}.
Differently, the HyperNet hardened by adversarial learning with nondeterministic adversarial examples shows better generalizable security against various types of adversarial examples.

$\blacksquare$ {\bf Adversary with HyperNet:}
A natural attack strategy for this kind of adversary is to follow the approach in \cite{liu2016delving} to craft adversarial examples against a {\em surrogate ensemble} generated by the HyperNet. We evaluate the SDRs of the dynamic ensembles generated by the same HyperNet.
Fig.~\ref{fig:sdr-vs-fpr-gtsrb-seperate-attack}e shows that the SDRs are much higher than those without MTD in which the adversary obtains the ensemble in use as mentioned at the beginning of this subsection (i.e., 100\% - 52\% = 48\% and 100\% - 37\% = 63\% when the FPR is 1.1\% and 5.2\%, respectively).
The SDR for the HyperNet with adversarial learning is higher than that without adversarial learning.

The above results suggest that the MTD of preventing the adversary from obtaining the ensemble in use is effective in counteracting adversarial example attacks. HyperNet-based MTD security is further enhanced with adversarial learning when the adversary constructs adversarial examples against the surrogates based on static information of the defense.

\subsubsection{Outlier detection performance}
\label{sec:outlier-detection}

\begin{figure}
  \centering
  \begin{minipage}[t]{0.485\columnwidth}
    \centering
    \includegraphics[width=\columnwidth]{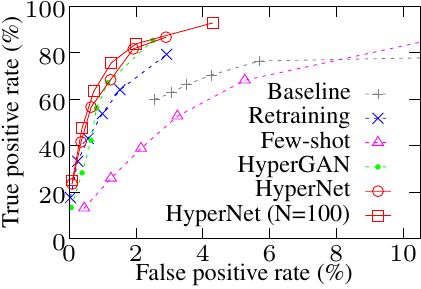}
    \vspace{-2em}
    \caption{Outlier detection (Dataset: MNIST \& notMNIST; $N=20$).}
    \label{fig:anomaly-detection}
  \end{minipage}
  \hfill
  \begin{minipage}[t]{.485\columnwidth}
    \centering
    \includegraphics[width=\columnwidth]{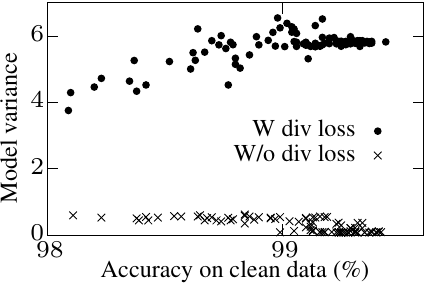}
    \vspace{-2em}
    \caption{Weights' variance of the ensemble DNNs (Dataset: MNIST).}
    \label{fig:acc-vs-var}
  \end{minipage}
  \vspace{-1em}
\end{figure}

We evaluate the outlier detection performance of the ensemble-based detectors and a baseline outlier detector described in \cite{hendrycks2016baseline}, which uses a single DNN and declares an outlier if the maximum of the softmax probabilities of all classes is below a threshold. We perform training using MNIST. During testing, we use notMNIST to assess the true positive rate of outlier detection and use MNIST to assess the false positive rate.
Fig.~\ref{fig:anomaly-detection} shows the receiver operating characteristic (ROC) of various outlier detectors. By default, $N=20$. To generate ROCs, we vary the consistency threshold $T_s$ from 50\% to 100\% for ensemble-based detectors and vary the softmax probability threshold from 90\% to 100\% for the baseline detector.
The HyperNet-ensemble's ROC curves for $N=20$ and $N=100$ are the highest in the plot, suggesting that HyperNet-ensemble outperforms other detectors. The ROC curve for $N=100$ is higher than that for $N=20$, suggesting that larger ensemble size is beneficial to outlier detection.

\subsubsection{Diversity of HyperNet-generated DNNs}

Fig.~\ref{fig:acc-vs-var} compares the HyperNets trained with or without the diversity loss $J_2$. Each point corresponds to a HyperNet-ensemble with $N=32$. The $x$-axis is the ensemble's accuracy on the MNIST clean samples. We calculate the variance for each weight parameter across all DNNs of an ensemble. The $y$-axis is the average of all weights' variances. We can see that the diversity loss $J_2$ diversifies the generated DNNs.

\subsection{Summary of Profiling Results}
\label{sec:observations}

From the profiling in \sect\ref{sec:measurement-results}, we can draw the following observations. First, HyperNet generates diverse DNNs that achieve high accuracy on clean examples. Second, HyperNet-ensemble outperforms adversarial training \cite{madry2018towards}, retraining-ensembles \cite{song2019moving,motiian2017few}, and HyperGAN-ensemble \cite{ratzlaff2019hypergan} in counteracting adaptive adversarial example attacks based on certain static information of the defense. Third, HyperNet-ensemble outperforms the OOD detection approaches based on softmax probability \cite{hendrycks2016baseline}, retraining-ensembles \cite{song2019moving,motiian2017few}, and HyperGAN-ensemble \cite{ratzlaff2019hypergan}. Lastly, HyperNet-ensemble's accuracy on clean examples and security/resilience against adversarial examples/outliers increase with $N$.
\section{Run-Time Planning of Ensemble Size}
\label{sec:ensemble-cardinality-scheduling}

From \sect\ref{sec:measurement-study}, it is desirable to maximize $N$ subject to the soft deadline of the resilient vision task. The key is the ability to predict the ensemble generation and execution time for any $N$ in the presence of time-varying background computation. The prediction should be compute-lightweight. With this ability, we can search the maximum $N$ meeting the deadline.

The general problem of scheduling GPU computing tasks to meet deadlines is challenging due to the non-preemptive nature of GPU kernels. 
Recent studies \cite{zhou2018s,zhang2018g} enable concurrent executions of multiple kernels and schedule the kernels to maximize processing throughput. 
Although solutions to the general problem are still lacking, our reduced problem of predicting the ensemble generation and execution time in the presence of uncoordinated background computation may have an effective solution if we can identify the major factors correlated with the ensemble generation/execution time and then apply supervised learning to characterize the correlations. Following this method, we conduct measurements to identify the correlated factors in \sect\ref{sec:impact-N}; we design and evaluate the ensemble latency predictor in \sect\ref{subsec:predictor} and \sect\ref{sec:model-eval}, respectively.

\subsection{Identifying Latency-Correlated Factors}
\label{sec:impact-N}

\begin{figure}
  \includegraphics[width=\columnwidth]{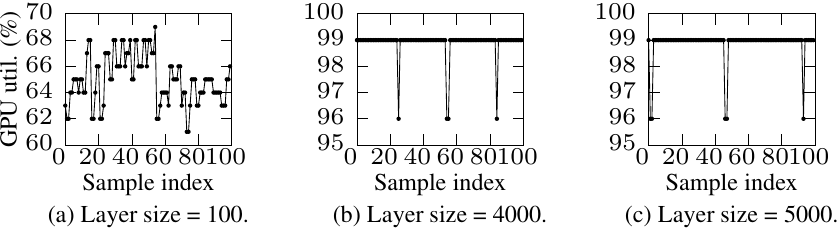} 
  \vspace{-2em}
  \caption{GPU utilization of background computation.}
  \label{fig:gpu-util-vs-time}
  \vspace{-1em}
\end{figure}

\begin{figure}
  \subfigure[Without ensemble]{
    \includegraphics[width=0.495\columnwidth]{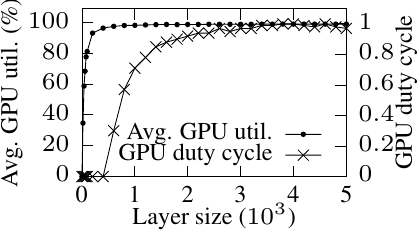}
    \label{fig:ls-vs-gpu-util-duty-cycle}
  }
  \hfill
  \subfigure[Without/with ensemble]{
    \includegraphics[width=.36\columnwidth]{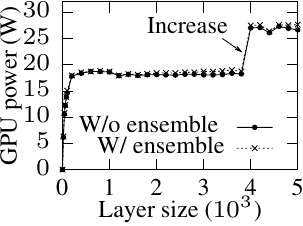}
    \label{fig:ls-vs-gpu-power-avg}
  }
  \vspace{-0.5em}
  \caption{Background GPU utilization and power usage vs. the layer size of the background DNN. ``W/o ensemble'' means only the background computation is running. ``W/ ensemble'' means the background computation and the ensemble generation and execution are running simultaneously.}
  \label{fig:back-gpu}
  \vspace{-0.5em}
\end{figure}

We set up a continuous DNN inference process as the {\em background computation}. 
As convolution is compute-intensive, we adjust the number of neurons of the background DNN's every convolutional layer (referred to as {\em layer size}) to affect the intensity of the background computation.
Fig.~\ref{fig:gpu-util-vs-time} shows the instantaneous GPU utilization traces on AGX Xavier when the layer size is 100, 4,000, and 5,000. Note that we use the \texttt{tegrastats} utility to measure the GPU utilization and power usage. When the layer size is 100, the GPU utilization fluctuates at 65\%. When the layer size is 4,000 and 5,000, the GPU utilization mostly remains at 99\%.
To understand the impact of the layer size on GPU utilization and power usage, we use the average value and duty cycle to characterize a GPU utilization trace obtained under a certain layer size. The duty cycle is the percentage of time at which the GPU utilization is higher than 99\%.
Fig.~\ref{fig:back-gpu} shows the GPU utilization and power when the layer size varies from 10 to 5,000. 
From Fig.~\ref{fig:ls-vs-gpu-util-duty-cycle}, GPU utilization's average and duty cycle increase smoothly with the layer size. The curve labeled ``Without ensemble'' in Fig.~\ref{fig:ls-vs-gpu-power-avg} shows a step increase of the GPU power when the layer size increases to 4,000. It can be caused by the increase of active stream processors to compute more neurons. The results in Fig.~\ref{fig:back-gpu} imply that GPU utilization and power depict different aspects of remaining GPU computing capability.

\begin{figure}
  \subfigure[]
  {
    \includegraphics[width=0.432\columnwidth]{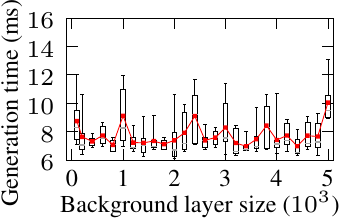}
    \label{fig:gen-time-vs-back-ls-error-bar}
  }
  \hfill
  \subfigure[]
  {
    \includegraphics[width=0.432\columnwidth]{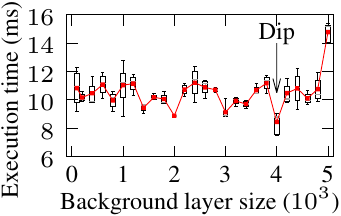}
    \label{fig:inf-time-vs-back-ls-error-bar}
  }
  \vspace{-0.5em}
  \caption{100-DNN ensemble generation/execution time vs. background computation layer size. (Grey line represents median; dot represents mean; box represents 20\%/80\% percentiles; whiskers represent max/min. Same style is applied for all error bars in this paper.)}
  \label{fig:time-vs-back-ls-error-bar}
  \vspace{-0.5em}
\end{figure}

\begin{figure}[t]
  \includegraphics[width=.95\columnwidth]{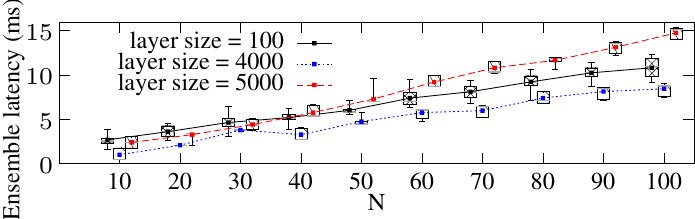}
  \vspace{-1em}
  \caption{Ensemble latency vs. ensemble size.}
  \label{fig:num-models-vs-inf-time-error-bar}
  \vspace{-1em}
\end{figure}
  
Fig.~\ref{fig:time-vs-back-ls-error-bar} shows how the background computation affects the delays for ensemble generation and execution. Both delays are relatively stable when the background layer size is up to 4,800. Both increase saliently when the background layer size increases from 4,800 to 5,000, which can be caused by the contention between background computation and ensemble generation/execution. In Fig.~\ref{fig:time-vs-back-ls-error-bar}, the ensemble execution time has a dip when the background layer size is 4,000. A potential reason is that, at this point, the GPU increases active stream processors as indicated by the sudden increase of GPU power in Fig.~\ref{fig:ls-vs-gpu-power-avg}. From Fig.~\ref{fig:time-vs-back-ls-error-bar}, it takes about $10\,\text{ms}$ to generate 100 DNNs using HyperNet. In contrast, the retraining approach in \cite{song2019moving} and the few-shot retraining approach in \cite{motiian2017few} require 45 and 22 minutes to generate 20 DNNs on AGX Xavier. Thus, HyperNet achieves 0.66 to 1.35 million times acceleration in per-DNN generation.

Fig.~\ref{fig:num-models-vs-inf-time-error-bar} shows the impact of $N$ on the ensemble execution time in the presence of background computation. Under a certain background layer size, the ensemble latency increases linearly with $N$ in general. When the layer size varies, the line of ensemble latency versus $N$ changes accordingly.

The above results show that the background GPU utilization and power, and $N$ are three factors correlated with the ensemble latency. From the near-linear relationships shown in Fig.~\ref{fig:num-models-vs-inf-time-error-bar}, simple models may effectively characterize the impact of these three factors on the ensemble latency.

\subsection{Design of Ensemble Latency Predictor}
\label{subsec:predictor}

\begin{figure}
  \includegraphics[width=\columnwidth]{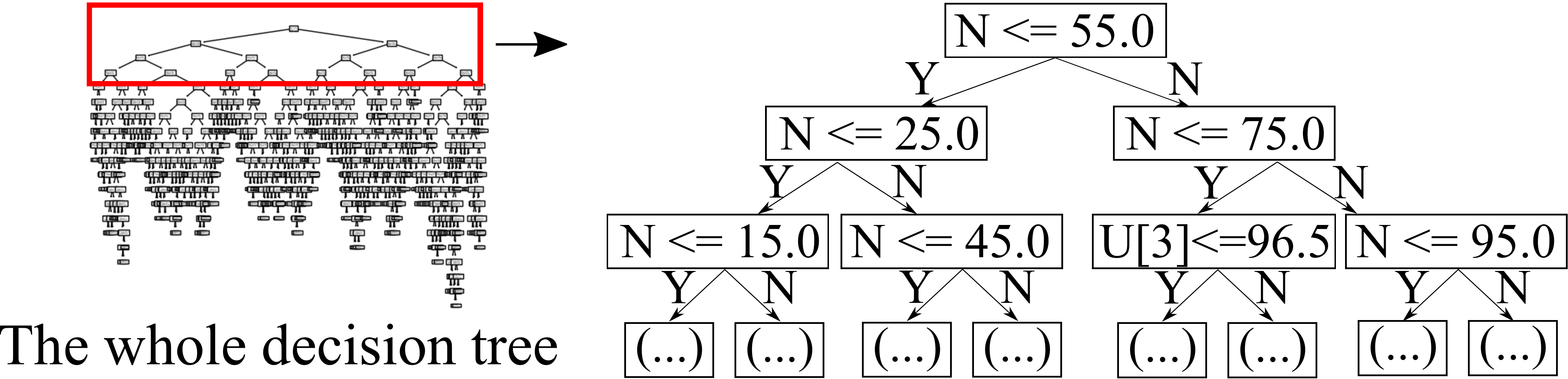}
  \vspace{-1.5em}
  \caption{Visualization of the decision tree for ensemble latency prediction on Jetson AGX Xavier. ``U[3]'' means the background GPU utilization collected at 30 ms before the start of the ensemble generation.}
  \label{fig:decision-tree-agx-cb}
  \vspace{-1em}
\end{figure}

We choose the background GPU utilization and power usage traces sampled by \texttt{tegrastats} at $100\,\text{Hz}$ in the past $100\,\text{ms}$ before the start of the ensemble generation and $N$ as the three inputs to the machine learning model. The output is the predicted ensemble latency.
We consider the following three candidate machine learning models and the evaluation in \sect\ref{sec:model-eval} and \sect\ref{sec:traffic-sign-recognition} will recommend a good choice.
(1) {\bf Decision tree (DT)} predicts the ensemble latency using a set of if-then-else decision rules. The tree structure and the decision rules associated with the tree nodes are learned from the training data. Fig.~\ref{fig:decision-tree-agx-cb} shows a DT and its top three layers learned from data collected from AGX Xavier. The root is the entry. Each node compares an element of the input with a threshold to decide which branch to proceed with. The tree leaves are the output nodes associated with predicted ensemble latency.
(2) {\bf Linear regression (LR)} predicts ensemble latency by a weighted sum of all inputs, where the weights are learned from the training data.
(3) {\bf DNN} uses two 200-neuron hidden layers with ReLU activation to predict the ensemble latency.

A trained predictor is specific to the mobile edge device's hardware and software configurations. The training needs a profiling process to collect training data. This overhead is acceptable since the profiling can be automated. Moreover,
the profiling and training are only performed by expert designers during system design and software updates.

Multi-core CPU shares some similarities with the many-core GPU in terms of hardware parallelism. Our design is also applicable to the CPU-only devices. Although CPU-only device is not suitable for real-time visual sensing and thus not our focus, we will briefly evaluate our design on CPU in \sect\ref{sec:traffic-sign-recognition}.

Lastly, we present how a trained predictor is used at run time. For each image input, Sardino queries the latest GPU utilization and power usage traces, predicts the ensemble latency for every candidate $N$ setting, and chooses the maximum setting that can meet the soft deadline. DT and LR predictors can be executed by CPU due to their low compute overheads.

\subsection{Evaluation of Latency Predictor}
\label{sec:model-eval}

\begin{figure}
  \subfigure[Jetson AGX Xavier]
  {
    \includegraphics[width=0.441\columnwidth]{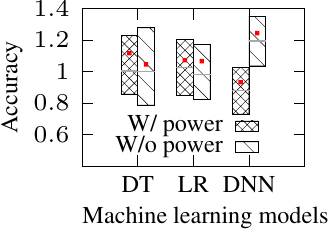}
    \label{fig:accuracy-error-bar-agx-cb}
  }
  \hfill
  \subfigure[Jetson Nano]
  {
    \includegraphics[width=0.441\columnwidth]{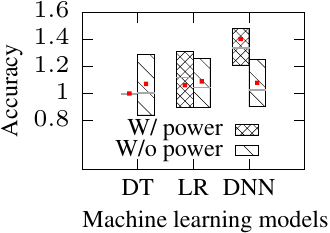}
    \label{fig:accuracy-error-bar-nano-cb}
  }
  \vspace{-0.5em}
  \caption{Ensemble latency prediction accuracy of the three models designed with or without GPU power usage trace as part of input. Accuracy is defined as the ratio between predicted value and true value.}
  \label{fog:accuracy-error-bar-cb}
  \vspace{-1em}
\end{figure}

We evaluate the prediction models described in \sect\ref{subsec:predictor}.
We collect 500 data points on AGX Xavier by varying the layer size of the background process from 100 to 5,000 with a step size of 100 and the ensemble size $N$ from 10 to 100 with a step size of 10.
We set the background process always running and start the ensemble generation and execution at random time instants. We shuffle and split the collected data into training and testing data with a ratio of 4:1.
The performance of the machine learning models is evaluated on the test data using two assessment metrics: (1) accuracy, defined as the ratio between the predicted and true values of ensemble latency, i.e., $\frac{T_{pred}}{T_{true}}$, and (2) root mean squared error (RMSE). The accuracies are shown in Fig.~\ref{fig:accuracy-error-bar-agx-cb}. The box plots labeled with ``W/ power'' are the results of the machine learning models with $N$, GPU utilization trace, and GPU power usage trace as input; those labeled with ``W/o power'' are the results of the models designed with $N$ and GPU utilization trace as input. Table~\ref{tab:ensemble-latency0} shows the three models' RMSEs. The inclusion of GPU power improves the prediction accuracy for DT and LR.

\begin{table}
  \caption{RMSE (ms) of ensemble latency prediction.}
  \vspace{-1em}
  \label{tab:ensemble-latency0}
  \centering
    \begin{tabular}{l|cc|cc}
    \hline
     & \multicolumn{2}{c|}{Jetson AGX Xavier} & \multicolumn{2}{c}{Jetson Nano} \\
          & W/ power & W/o power & W/ power & W/o power \\
    \hline
    DT & 1.59 & 1.61 & 2.60e-6 & 16.70 \\
    LR & 1.09 & 1.18 & 14.60 & 23.31 \\
    DNN & 1.41 & 1.36 & 15.58 & 16.19 \\
    \hline
    \end{tabular}
    \vspace{-1em}
\end{table}

Then, we evaluate whether the machine learning approach works for NVIDIA Jetson Nano, which is a less powerful platform with a quad-core Cortex-A57 CPU, a 128-core Maxwell GPU, and 4GB RAM. 
The results are shown in Fig.~\ref{fig:accuracy-error-bar-nano-cb} and Table~\ref{tab:ensemble-latency0}. 
The inclusion of GPU power into input also improves prediction accuracy. DT outperforms the other two models.

From Fig.~\ref{fog:accuracy-error-bar-cb} and Table~\ref{tab:ensemble-latency0},
the three prediction models achieve similar accuracy on AGX Xavier; the DT outperforms significantly on Nano. 
In terms of compute overhead, DT's time complexity is linear to its depth, which is sub-linear to the number of decision variables. Compared with LR and DNN that have linear and super-linear time complexities, DT is more efficient and preferred. 
On both Jetson boards, DT achieves sub-$2\,\text{ms}$ RMSEs.
This accuracy is acceptable for meeting the soft deadlines of tens of milliseconds.
The prediction errors will cause jitters, which will be evaluated in \sect\ref{sec:traffic-sign-recognition}.

DT's superior performance is because the hierarchical structure of DT can better capture the priority hierarchy of the affecting factors in determining the ensemble latency. For instance, in Fig.~\ref{fig:decision-tree-agx-cb}, the top layers of the tree make decisions based on $N$. The conditions for the latest GPU utilization appear at lower layers. These match with the observations from Fig.~\ref{fig:num-models-vs-inf-time-error-bar} that (1) $N$ determines the range of the ensemble latency value and (2) the background computation intensity determines which line to follow and the exact value. In contrast, LR is short of capturing such non-linear priority hierarchy. 
Although DNN can capture sophisticated patterns, it needs a rich training dataset. 
A possible reason for DNN's degraded accuracy on Nano than AGX Xavier is that the more GPU contention on the less powerful Nano increases the pattern complexity and thus requires more training data.

Note that other factors such as environment temperature and processor cache hit rate may also generate impact on the ensemble latency. Including these factors to the machine learning model's input may further improve prediction accuracy. We leave this further improvement to future study.
\section{Real-Time On-Car Traffic Sign Recognition}
\label{sec:traffic-sign-recognition}

\subsection{System Implementation}
\label{sec:system-impl}

\begin{figure}[t]
  \centering
  \includegraphics[width=\columnwidth]{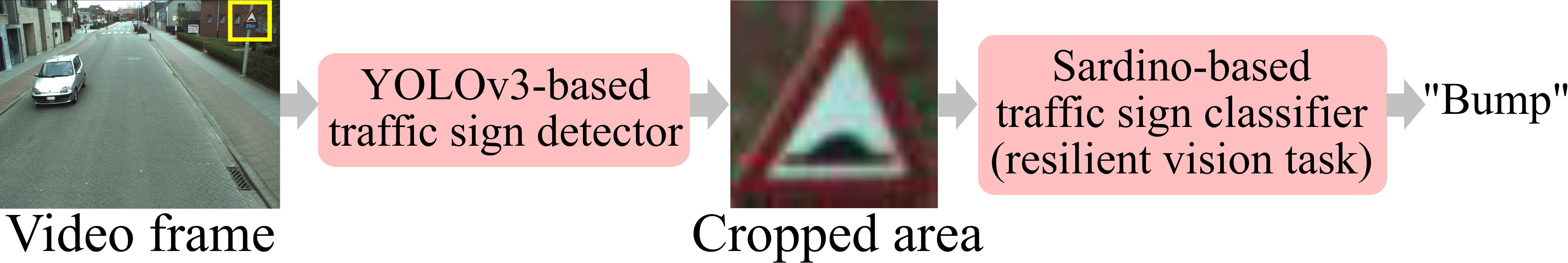}
  \vspace{-2em}
  \caption{The pipeline of the traffic sign recognition.}
  \label{fig:sign-recog-pipline}
  \vspace{-1em}
\end{figure}

We apply Sardino to build a real-time car-borne traffic sign recognition system.
A traffic sign recognition system usually consists of the sign detection and classification phases \cite{zhu2016traffic2}. The detector identifies and locates traffic signs in an incoming frame captured by a car-mounted camera. The detected traffic sign is then interpreted by the classifier.
We implement a Belgian traffic sign recognition system based on the publicly available KUL Belgium Traffic Signs Dataset \cite{timofte2014multi}, which has two datasets for traffic sign detection (BelgiumTSD) and classification (BelgiumTSC) and four recorded videos. 

Fig.~\ref{fig:sign-recog-pipline} illustrates the processing pipeline of our implementation.
We extend YOLOv3 and YOLOv3-tiny \cite{yolo} as the traffic sign detector.
YOLOv3 is a DNN-based object detection system that achieves good accuracy and latency performance among various systems and YOLOv3-tiny is a simplified version with fewer layers and same class number.
We use BelgiumTSD to augment the original training set for YOLOv3 and YOLOv3-tiny and retrain them to achieve mAP@0.5 of 58.9 and 33.7, respectively. Note that mAP@0.5 is the mean Average Precision when a prediction is considered positive if intersection over union (IoU) is no smaller than 0.5. The mAP@0.5 scores for the original YOLOv3 and YOLOv3-tiny are 57.9 and 33.1 \cite{yolo}.
The extended YOLOv3 and YOLOv3-tiny can detect the Belgian traffic signs as the class ``traffic sign'' and use a bounding box to contain each detected sign. The bounding box area is cropped from the original frame, resized, and passed to the Sardino-based traffic sign classifier. Other objects detected are not processed.
We train Sardino's HyperNet using the 62-class BelgiumTSC training set. The HyperNet-generated DNNs achieve an average accuracy of 96.1\% on the BelgiumTSC testing set, which is comparable to the accuracy of 97.0\% reported in \cite{timofte2014multi}. We deploy the extended YOLOv3 and YOLOv3-tiny on Jetson AGX Xavier and Jetson Nano, respectively. On AGX Xavier, YOLOv3's processing throughputs are 82, 55, and 42 fps when the input frame sizes are 320x320, 416x416, and 512x512, respectively. On Nano, YOLOv3-tiny's processing throughputs are 37, 26, and 22 fps for the three input frame sizes.

\subsection{Performance Evaluation}
{\blue 
}

\subsubsection{Ensemble latency prediction}
\label{sec:ensemble-latency-gpu}

\begin{figure}
  \subfigure[AGX Xavier, YOLOv3]
  {
    \includegraphics[width=0.4165\columnwidth]{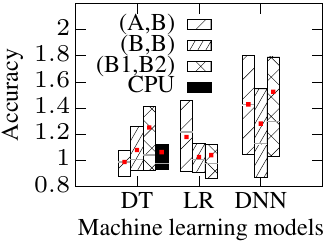}
    \label{fig:accuracy-error-bar-agx-Yolo}
  }
  \hfill
  \subfigure[Nano, YOLOv3-tiny]
  {
    \includegraphics[width=0.4165\columnwidth]{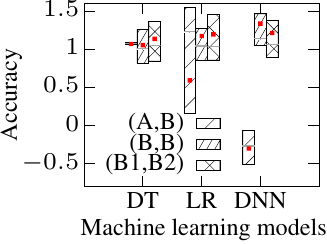}
    \label{fig:accuracy-error-bar-nano-Yolo}
  }
  \vspace{-0.5em}
  \caption{Ensemble latency prediction on car-borne traffic sign recognition.}
  \label{fig:accuracy-error-bar-Yolo}
  \vspace{-1em}
\end{figure}

\begin{table}
  \caption{RMSE (ms) of ensemble latency prediction.}
  \vspace{-1em}
  \label{tab:rmse-yolo}
  \centering
  \begin{tabular}{l|ccc|ccc}
    \hline
    Board & \multicolumn{3}{c|}{Jetson AGX Xavier} & \multicolumn{3}{c}{Jetson Nano} \\
    (train,test) & (A,B) & (B,B) & (B1,B2) & (A,B) & (B,B) & (B1,B2) \\
    \hline
    DT & \underline{1.73} & 1.18 & 1.25 & \underline{1.49} & 11.01 & \underline{11.64} \\
    LR & 1.77 & \underline{0.69} & \underline{0.88} & 14.08 & \underline{8.09} & 12.75 \\
    DNN & 2.91 & 1.96 & 2.82 & 4.52 & 8.47 & 20.23 \\
    \hline
    \multicolumn{7}{l}{*The minimum RMSE among the three models is underlined.}
  \end{tabular}
  \vspace{-1em}
\end{table}

The YOLOv3 on the AGX Xavier and YOLOv3-tiny on Nano are viewed as the background computation.
We set the frame sizes to be 320x320, 416x416, and 512x512 to obtain different background computation intensities. For each frame size, we vary $N$ from 10 to 100 with a step size of 10. We repeat each setting for 20 times and collect 600 data points for evaluation on each platform. Fig.~\ref{fig:accuracy-error-bar-Yolo} shows the ensemble latency prediction accuracy and Table~\ref{tab:rmse-yolo} shows RMSEs. In particular, we evaluate the transferability of the trained prediction models across different background computations. Label (A,B) means the model is trained with the customized background computation described in \sect\ref{sec:impact-N} and tested using the YOLO background computation with all frame size settings; label (B,B) means the model is trained and tested using the YOLO background computation; label (B1,B2) means the model is trained using the YOLO background computation with 320x320 and 512x512 frame sizes and tested with 416x416 frame size. From Table~\ref{tab:rmse-yolo}, DT and LR achieve similar RMSEs. However, on Nano with setting (A,B), LR performs poorly. The results also show that DT exhibits good transferability across different background computations. In the rest of this paper, we use DT.

To evaluate whether DT is applicable to CPU-only devices, we run YOLO background computation and ensemble generation/execution on the CPU of AGX Xavier. The inputs to DT are $N$, background CPU utilization, and CPU power usage. The DT is trained using data traces when the system operates on 320x320 and 512x512 frame sizes and tested on 416x416 frame size.
The prediction accuracy, as shown in Fig.~\ref{fig:accuracy-error-bar-agx-Yolo} by the box labeled ``CPU'', is comparable to those on GPU. However, YOLO (without Sardino) only achieves a throughput of $0.05\,\text{fps}$ on CPU.
Thus, CPU-only devices are ill-suited for real-time visual sensing although the DT is still applicable.

\subsubsection{Real-time performance of Sardino}
\label{sec:real-time-video}

\begin{figure}
  \centering
  \includegraphics[width=.95\columnwidth]{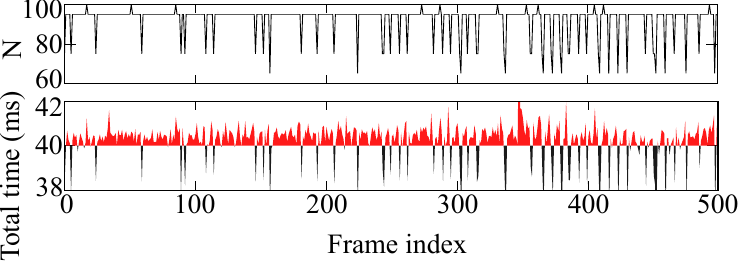}
  \vspace{-1em}
  \caption{Traces of planned $N$ and per-frame total processing time. Frame rate: $25\,\text{fps}$; deadline: $40\,\text{ms}$.}
  \label{fig:CDF-FPS-AGX}
  \vspace{-1em}
\end{figure}

\begin{figure}
  \subfigure[Deadline: 50ms]
  {
    \includegraphics[width=.295\columnwidth]{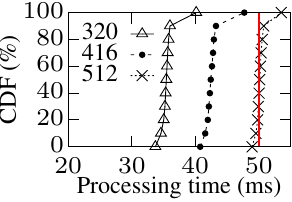}
    \label{fig:Yolo-fp20-CDF}
  }
  \subfigure[Deadline: 40ms]
  {
    \includegraphics[width=.295\columnwidth]{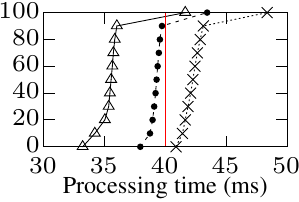}
    \label{fig:Yolo-fps25-CDF}
  }
  \subfigure[Deadline: 33ms]
  {
    \includegraphics[width=.295\columnwidth]{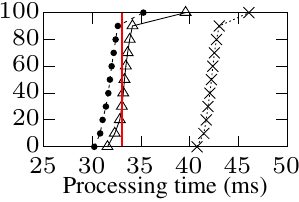}
    \label{fig:Yolo-fps30-CDF}
  }
  \vspace{-1.5em}
  \caption{CDF of per-frame processing time under three deadline settings (represented by vertical lines) and three frame size settings of $320 \times 320$, $416 \times 416$, and $512 \times 512$ (different CDFs for different frame sizes).}
  \label{fig:CDF-proc-time-AGX}
  \vspace{-1em}
\end{figure}

The total processing time for each frame consists of: (1) YOLOv3 detection time, which depends on the frame size; (2) ensemble generation time; and (3) ensemble execution time. YOLOv3's detection time is measured at run time. Then, we follow the method presented in \sect\ref{subsec:system-model} to determine the soft deadline and pass it to the ensemble size planner.
Fig.~\ref{fig:CDF-FPS-AGX} shows the planned $N$ and per-frame processing time traces, where the target per-frame processing time is $40\,\text{ms}$.
We repeatedly play a video to introduce disturbances during the experiments. 
We can see that Sardino frequently adjusts $N$. The per-frame processing time fluctuates at $40\,\text{ms}$. 
The average per-frame processing time is $40.43\,\text{ms}$; the maximum deviation is about $2\,\text{ms}$, consistent with the error level of the DT predictor. 

Fig.~\ref{fig:CDF-proc-time-AGX} shows the CDFs of the per-frame processing time on AGX Xavier under various settings of deadline and YOLOv3 frame size. When the specified deadline is $50\,\text{ms}$ for processing a $20\,\text{fps}$ frame stream, from Fig.~\ref{fig:Yolo-fp20-CDF}, the deadline can be always met for frame sizes of 320x320 and 416x416, and mostly met for frame size of 512x512. For the former two cases, the allowed time of $50\,\text{ms}$ is not fully utilized, because we set an upper bound of 100 for $N$. Larger settings for $N$ often lead to memory exhaustion. 
When the specified deadline is $40\,\text{ms}$, the system with 320x320 and 416x416 frame sizes can still largely meet the deadline. When the frame size is 512x512, YOLOv3's detection time is very close to the $40\,\text{ms}$ deadline. The deadline may be exceeded even if the minimum setting $N = 3$ is chosen.
When the specified deadline is $33\,\text{ms}$, the system with 512x512 frame size completely misses the deadline because the uncontrollable YOLOv3 detection time already exceeds the deadline. The system with the other two frame sizes can still largely meet the deadline. Thus, by properly choosing settings that will not overwhelm the system, Sardino can maximize $N$ while meeting required frame rate.

\subsubsection{Resilience against OOD data}
\label{sec:sensing-resilience}

\begin{figure}
  \centering
  \includegraphics[width=\columnwidth]{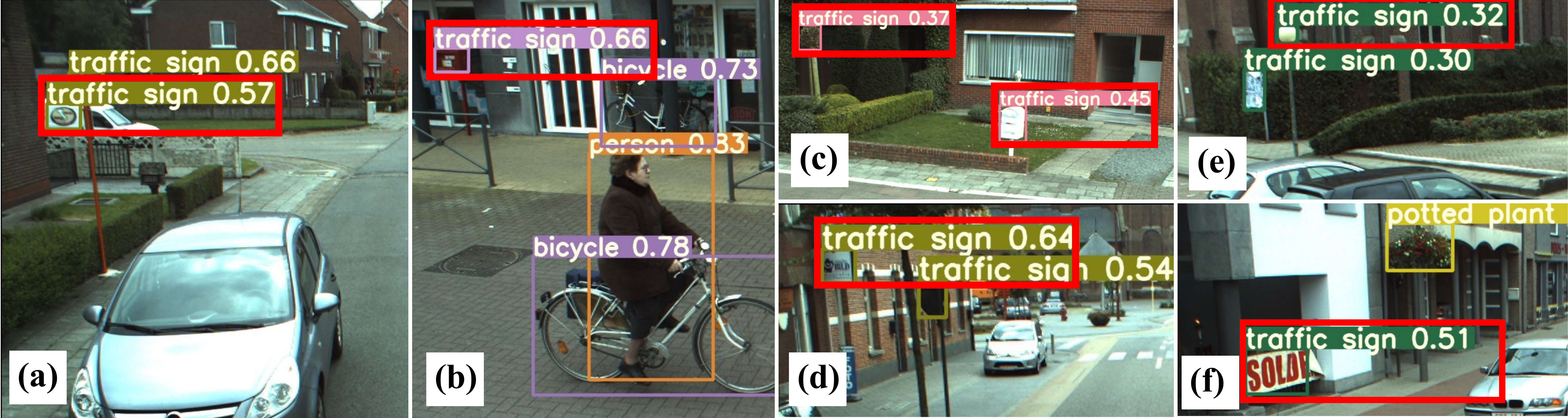}
  \vspace{-2em}
  \caption{Some OOD samples (highlighted by red frames) caused by YOLO's false positives in detecting traffic sign. (a) pattern on a car; (b) mailbox; (c) plant \& mailbox; (d) shop signboard; (e) street light; (f) banner.}
  \label{fig:outliers-pic}
\end{figure}

\begin{figure}
  \begin{minipage}{0.46\columnwidth}
    \centering
    \includegraphics[width=.78\columnwidth]{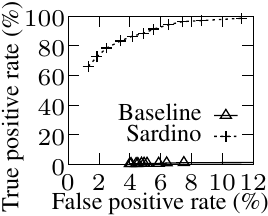}
      \vspace{-1em}
      \caption{Outlier detection performance of Sardino.}
      \label{fig:outlier-detection-yolo}
    \end{minipage}
  \hfill
  \begin{minipage}{.48\columnwidth}
    \centering
    \includegraphics[width=.78\columnwidth]{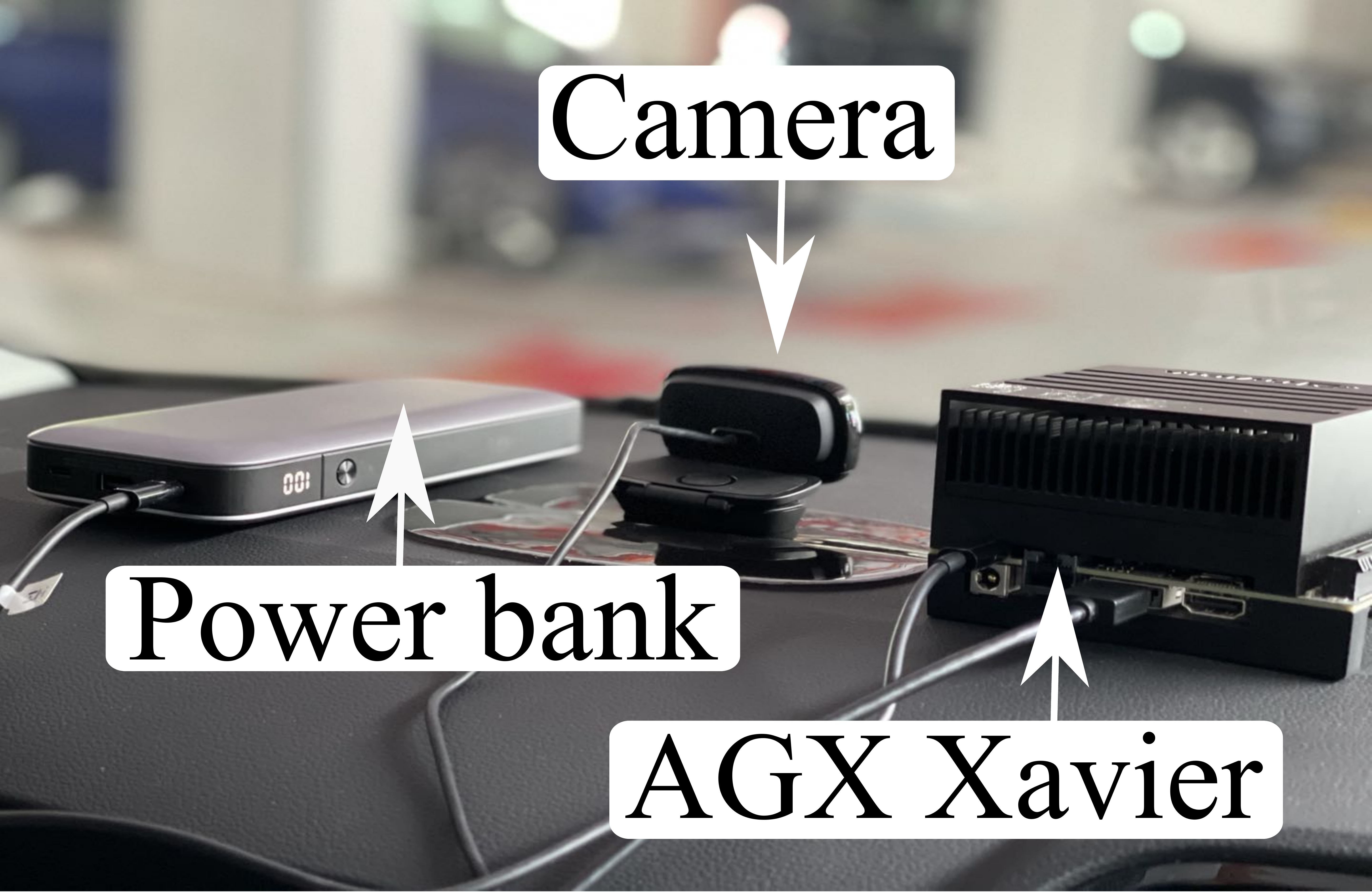}
    \caption{Setup installed under a car's front windshield.}
    \label{fig:driving-test-setup}
  \end{minipage}
  \vspace{-0.5em}
\end{figure}

The false positives of the YOLO-based traffic sign detector are naturally occurring OOD data for traffic sign classifier. Fig.~\ref{fig:outliers-pic} shows six examples of such false positives.
We measure Sardino's true positive rate in detecting OOD data using 500 YOLO's false positives in processing the four test videos of the KUL dataset. Fig.~\ref{fig:outlier-detection-yolo} shows the ROCs of the Sardino and the baseline detector described in \sect\ref{sec:outlier-detection}. The baseline detector is ineffective. This is because YOLO's internal detector only yields objects (including false positives) detected with high confidence.
Fig.~\ref{fig:outlier-detection-yolo} shows that Sardino advances the resilience of traffic sign classification against YOLO's false positives.

We do not measure Sardino's performance in thwarting adversarial examples in this application. This is because, the numeric experiment results in \sect\ref{sec:thwarting-adversarial} based on real traffic sign data and ideal attack settings (e.g., pixel-level perturbation capability) characterize the lower bound of Sardino's attack thwarting performance. The results in \sect\ref{sec:thwarting-adversarial} have already shown the superior performance of Sardino. Differently, the numeric experiments in \sect\ref{sec:outlier-detection} on OOD detection are based on the simplistic handwritten digit recognition task for illustration only. Thus, in this section, we focus on evaluating Sardino's OOD detection performance for this real-world application of traffic sign recognition.

\subsubsection{Stress tests on live roads}
\label{sec:driving-test}

\begin{figure}
  \centering
  \includegraphics[width=\columnwidth]{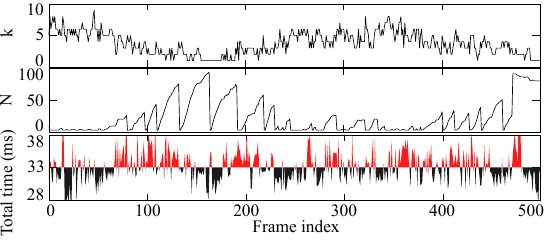}
  \vspace{-2em}
  \caption{Number of detected objects ($k$), planned $N$, and per-frame total processing time under setting \ding{182}.}
  \label{fig:driving-test}
  \vspace{-1em}
\end{figure}

The test videos in the KUL dataset have a limited number of objects in the camera's field of view. To evaluate the performance of Sardino under more challenging settings, we conduct live tests on the busy roads of Singapore. To stress-test Sardino's real-time performance, we let Sardino process each object detected by YOLO, not limited to traffic sign objects. As shown in Fig.~\ref{fig:driving-test-setup}, we install our system under a car's front windshield. The AGX Xavier is connected with a Logitech C525 camera via USB and powered by a portable battery. We drove the car for multiple runs, each lasting for about 30 minutes. Each run covers various road types, including campus, locals, and expressways. We test three video settings in terms of frame size and rate: \ding{182} 320x320 at $30\,\text{fps}$; \ding{183} 416x416 at $24\,\text{fps}$; \ding{184} 512x512 at $20\,\text{fps}$.

Fig.~\ref{fig:driving-test} shows the number of detected objects $k$, the planned $N$, and the per-frame total processing time in about 17 seconds during a run. The number of detected objects in a frame can be up to 9. Sardino frequently adjusts $N$ to maintain the per-frame processing time at $33\,\text{ms}$. The average per-frame processing times under the three video settings are $33.0\,\text{ms}$, $40.4\,\text{ms}$, and $50.5\,\text{ms}$, respectively. Although the system has jitters, the average processing times are very close to the setpoints of $33\,\text{ms}$, $41.7\,\text{ms}$, and $50\,\text{ms}$.

\section{Discussions}
\label{sec:discuss}

A possible concern is that the ensemble's mutability may impede post-incident faulty analysis in the context of autonomous driving,
because storing each renewed ensemble incurs high storage overhead.
In fact, we only need to store the random seeds fed to HyperNet, which introduces only 5.5MB/hour storage overhead. With the seeds, we can trace back to the ensembles in fault analysis. Another related concern is that the inference accuracy of the renewed ensembles is not validated. To mitigate this concern, a validation dataset can be used to test the inference accuracy of each renewed ensemble at run time. Only the ensembles passing the test will be commissioned. On Jetson AGX Xavier, it takes about 7.2 seconds to complete the validation of a 100-DNN ensemble using 4,410 samples. Therefore, with this validation process, Sardino cannot achieve the per-frame ensemble renewal. However, compared with the off-time retraining-based approaches \cite{song2019moving,motiian2017few}, the run-time ensemble renewal at a rate of every 7.2 seconds provides much stronger MTD security and avoids battery over-discharge as discussed in \sect\ref{sec:intro}.

In the traffic sign recognition application, the traffic sign detector can be also vulnerable to adversarial example attacks. Sardino can be easily extended to support multiple pipelined resilient vision tasks (e.g., traffic sign detector and classifier). Specifically, for each image frame, we can allocate the remaining processing time of $\frac{1/x - t_d}{k}$ calculated using the approach described in \sect\ref{subsec:system-model} to the multiple resilient vision tasks by following a pre-defined policy (e.g., equal split) and use the respective ensemble size planner for each resilient vision task.
\section{Conclusion}
\label{sec:conclusion}

This paper presented Sardino, a HyperNet-based ultra-fast MTD approach for visual sensing at edge. Sardino generates quality ensembles that provide good classification accuracy on clean data and improved resilience against adversarial examples and naturally occurring OOD inputs. With the ultra-fast ensemble renewal and ensemble generation/execution time prediction, Sardino continuously updates the ensemble size such that each video frame can be processed with a new ensemble within a soft deadline, rendering the highest level of MTD security against adaptive adversarial example attacks. We use Sardino to build a real-time car-borne traffic sign recognition system and extensively evaluate its performance.

\bibliographystyle{plain}
\bibliography{reference}

\end{document}